\documentclass{article} 
\usepackage{iclr2024_conference,times}

\usepackage{hyperref}
\usepackage{url}
\usepackage{fancyhdr}

\fancypagestyle{firstpage}{
    \fancyhf{} 
    \fancyhead[L]{Preprint}
    \fancyfoot[L]{\textcopyright \ 2024 Christian A. Schiller.}

}

\usepackage{algorithm}
\usepackage{algorithmic}
\usepackage{amsthm}
\usepackage{caption}
\usepackage{subcaption}
\usepackage{mathtools}
\usepackage{adjustbox}
\usepackage{multirow}
\usepackage{siunitx}

\usepackage{enumitem}
\setlist{leftmargin=3mm}

\usepackage{wrapfig,lipsum,booktabs}
\usepackage{needspace}

\usepackage{longtable}
\usepackage{multirow}
\usepackage{array}

\title{The Human Factor in Detecting Errors of Large Language Models: A Systematic Literature Review and Future Research Directions}

\author{%
    Christian A. Schiller \\
    University of Duisburg-Essen, 45141 Essen, Germany \\
    christian.schiller@stud.uni-due.de \\
}

\iclrfinalcopy 
\begin{document}

\thispagestyle{firstpage} 

\maketitle

\begin{abstract}
The launch of ChatGPT by OpenAI in November 2022 marked a pivotal moment for Artificial Intelligence, introducing Large Language Models (LLMs) to the mainstream and setting new records in user adoption. LLMs, particularly ChatGPT, trained on extensive internet data, demonstrate remarkable conversational capabilities across various domains, suggesting a significant impact on the workforce. However, these models are susceptible to errors - ``hallucinations'' and omissions, generating incorrect or incomplete information. This poses risks especially in contexts where accuracy is crucial, such as legal compliance, medicine or fine-grained process frameworks.

There are both technical and human solutions to cope with this isse. This paper explores the human factors that enable users to detect errors in LLM outputs, a critical component in mitigating risks associated with their use in professional settings. Understanding these factors is essential for organizations aiming to leverage LLM technology efficiently, guiding targeted training and deployment strategies to enhance error detection by users. This approach not only aims to optimize the use of LLMs but also to prevent potential downstream issues stemming from reliance on inaccurate model responses. The research emphasizes the balance between technological advancement and human insight in maximizing the benefits of LLMs while minimizing the risks, particularly in areas where precision is paramount.

This paper performs a systematic literature research on this research topic, analyses and synthesizes the findings, and outlines future research directions. Literature selection cut-off date is January 11th 2024.
\\ \\
\textbf{Keywords}: generative artificial intelligence, genai, generative pretrained transformer, gpt, large language model, llm, llm error, llm hallucination, gpt error, gpt hallucination
\end{abstract}

\section{Introduction}

The release of the ChatGPT service by OpenAI in November 2022 started a new hype cycle for artificial intelligence (Mellick, 2022). The release of ChatGPT made the LLM technology popular in wider society and set a record for the fastest-growing user base in the history of the internet, reaching 100 million monthly active users just two months after launch (Hu, 2023). ``Generative Pretrained Transformer'' (GPT) is a new class of machine learning (ML) model initially developed by Google in 2017 (Vaswani et al, 2017). When trained with large text data bases, GPT-based ML models are also called “Large Language Model” (LLM). When queried by users with text inputs (“prompts”), LLMs can generate natural language responses. The ChatGPT LLM was trained on a large corpus of crawlable internet data. It exhibits human-level conversational skills and stores an unprecedented breadth of knowledge from its large training data base. For any question or task from any domain of human knowledge posed to ChatGPT, ChatGPT produces responses of – seemingly – high quality. Prompts such as “write a birthday invitation in the form of a Shakespearean sonnet” cater to tasks from personal life, whereas prompts like “which privacy requirements do the EU data privacy laws mandate for personal data” cater to tasks in professional life.
This type of professional prompting of LLMs is making quick inroads into everyday work in almost all human non-physical labour professions. Around 80\% of the U.S. workforce could have at least 10\% of their work tasks affected by the introduction of LLMs, while approximately 19\% of workers may see at least 50\% of their tasks impacted (Elounou et al, 2023).
However, the seemingly high quality of ChatGPT responses is not as high as it seems at first glance. LLM systems are a variant of deep neural networks.  Based on the input prompt, they simply predict each response word sequentially, choosing the statistically most likely response based on their training data corpus (Maynez et al, 2020). Due to the inherent stochastic nature of this process, deep learning-based text generation is prone to 'hallucinating' unintended text (Maynez et al, 2020). A typical example of such hallucinations: Having prompted ChatGPT to provide literature references for a previously discussed topic, ChatGPT produced five references which look plausible given the topic, but three of them turned out to be false (Emsley, 2023). Relating back to our professional prompting example of gathering EU data privacy requirements, it becomes clear why blindly trusting LLM responses could be costly, assuming that some of the requirements in the ChatGPT response are hallucinated: Businesses can be fined with up to 10\% of their yearly revenue when infringing EU data privacy laws, which can happen when mandated requirements are implemented insufficiently.
There are technological solutions to the hallucination problem which the LLM system developers are continually working on. Hallucination root causes such as noisy training data, imperfect representation learning or erroneous decoding can be addressed in the LLM system’s development process (Ji et al, 2023). But there is also a human solution to the problem: When a professional user can detect such LLM hallucinations, the user can prevent downstream problems of using a hallucinated LLM response for further work right from the start. The more LLM systems are used in the professional domain, the more potential downstream problems could occur – and prevented, if hallucinations do not occur (technical solution) or are spotted by the user (human solution).
The rollout of LLM systems is costly. License costs of commercial LLMs such as OpenAI ChatGPT are \$20-\$50 per user per month, which can quickly sum up to millions of dollars per month in larger organisations. Knowing the human factors which contribute to effective use of LLM systems allows organisations to choose a cost-efficient targeted rollout of LLMs to matching groups of users, instead of performing a costly full rollout of LLMs for all users. It also allows for preventing downstream problems arising from ``blind use'' of LLM responses, by training users appropriately in those human factors that are known to increase likelihood of spotting hallucinations (LLM errors). This leads to the research question in scope of this work:

\emph{}%
\noindent\fbox{\begin{minipage}[t]{1\columnwidth - 2\fboxsep - 2\fboxrule}%
\emph{Which human factors influence the ability of humans to detect LLM system errors?}%
\end{minipage}}

When the human factors in the ability to spot LLM errors are known, organisations can better select and more appropriately train professional users to increase the effectiveness of using LLM systems. Depending on the identified factors, this could inspire further research, such as: How well are LLM hallucinations spotted related to the seniority of a user – should rollout start from senior users first to junior users later, or vice versa? What impact would there be if junior users would fall for hallucinations (i.e. not detect them) and thus cause grave errors in downstream tasks relying on the hallucinated LLM responses? 

The objective of this paper is to summarize the current state of knowledge on the research question using a systematic literature review. For this, the framework defined by vom Brocke et al (2009) is used, which is summarised in chapter 2.

\section{Methodology}

This literature review follows a systematic framework for literature review consisting of five phases, with the goal to improve the rigour (validity, reliability) of the literature review process (vom Brocke et al, 2009). Chapters 3 to 7 of this document correspond to phases I to V of this review process.

\begin{center}
\begin{figure}[!h]
\begin{centering}
\includegraphics[width=11cm]{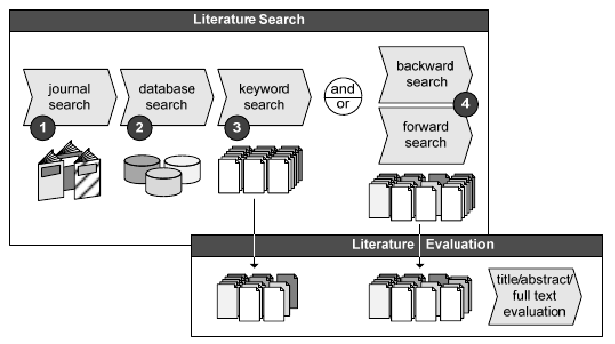}
\par\end{centering}
\caption{Framework for systematic literature research (vom Brocke et al, 2009)}
\end{figure}
\par\end{center}

Chapter 3 / Phase I further defines the research purpose and scope using the recommended methods. Chapter 4 / Phase II conceptualises the topics of LLM hallucinations and human factors in detecting them as a basis to inform the literature search. Chapter 5 / Phase III selects the literature search sources and defines literature search keywords based on the research scope and conceptualised topic. It presents the resulting list of literature to be analysed. Chapter 6 / Phase IV analyses the literature using a concept matrix and synthesizes the existing knowledge for each concept across the selected literature. Chapter 6 / Phase V discusses the synthesized findings considering the research question and identifies research gaps to inform a research agenda.

\section{Research scope}

The purpose of this review positioned in the Hart (1998) taxonomy is to synthesise existing research outcomes and research methods. Although focusing on the research outcomes already satisfies the research question and informs research gaps, additionally focusing on the research methods will inform with which methods further research on the research gaps can be performed, e.g. which human factor categorization schemes have been applied (if any).

Positioning this review in Cooper (1988) taxonomy is summarised in table 1. The focus of the review is on research outcomes (``Which human factors contribute to identifying LLM errors?'') and research methods (``How is this currently researched?''). The research goal is integration of existing knowledge. Therefore, a neutral representation is strived for. Audience are specialised scholars in the domains of information systems and business informatics at the intersection with data science and machine learning domains. The coverage of literature is only representative not exhaustive, because the literature review was performed within tight constraints of duration (3 months) and effort (part-time work).

\begin{table}[!h]
\centering
\begin{tabular}{lp{10cm}}
\toprule
\textbf{Characteristic} & \textbf{Categories (bold typeface: in scope of review)} \\
\midrule
Focus & \textbf{Research outcomes}, \textbf{Research methods}, Theories, Applications \\
Goal & \textbf{Integration}, \textbf{Criticism}, Central Issues \\
Organisation & historical, \textbf{conceptual}, methodological) \\
Perspective & \textbf{Neutral representation}, Espousal of position \\
Audience & \textbf{Specialised scholars}, General scholars, Practitioners/Politicians, General public \\
Coverage & Exhaustive, Exhaustive and selective, \textbf{Representative}, Central/Pivotal \\
\bottomrule
\end{tabular}
\caption{Positioning of this literature review in the taxonomy of Cooper (1988)}
\label{table1}
\end{table}

\section{Conceptualisation of topics}

The following working definitions of the key terms relating to the research question are introduced: ``LLM error'' and ``Human factors in detecting an LLM error''.

\subsection{LLM error}

An LLM error occurs when an LLM system (such as ChatGPT) provides an erroneous response to a given question or task. For this, machine learning research adopted the term ``hallucination'' (Maynez et al, 2020) from the psychology domain and transferred it to LLM systems as ``generated content that is nonsensical or unfaithful to the provided source content'' (Ji et al, 2023).

From the perspective of factuality, such LLM errors can be of two types:
\begin{enumerate}
    \item The LLM response contains factually incorrect information. For example, when the LLM system responds ``Italy'' when asked ``Who initiated World War II?'', while the correct response should be ``Germany''. Or when an LLM system responds ``Beef, Milk, Cheese, Eggs'' when asked ``What are agrarian products derived from cows?'', while ``Eggs'' should not be part of the response.
    \item The LLM response omits factually correct information even though it is relevant for the question or task prompted. For example, when the LLM system responds ``Physical, Network, Application'' when asked ``What are all layer names of the OSI model?'', while the correct response should include all seven OSI model layer names.
\end{enumerate}

From the perspective of the root cause, such LLM errors can be distinguished in two types, intrinsic, the error is caused by erroneous training data, or extrinsic, the error cause can not be found in the training data (Ji et al, 2023). From this definition, it becomes clear that hallucinations are in some cases even wanted/desirable behaviour of an LLM system, for example in the case of poem or story writing or any other artistic tasks. For other tasks however, such as correctly listing all EU data privacy law requirements, these errors are unwanted. This is in focus of this literature review.

\subsection{Human factors in detecting an LLM error}

For this paper, it is defined as ``a \textbf{personal attribute} or \textbf{personality trait} a human LLM user possesses that may make him or her more or less likely to identify an LLM error.''
Regarding \textbf{personal attributes}, these could for example be demographic information (age, gender etc.), living or work location, educational background, familiarity with the subject matter, level and/or duration of professional experience, or general technical proficiency (digital literacy). The framework to be used here is allowed personal data according to GDPR (Mondschein et al, 2019). This includes the attribute examples above, but excludes sensitive personal data such as (i) racial or ethnic origin, (ii) political opinions, (iii) religious or philosophical beliefs, (iv) trade union membership, (v) genetic data, (vi) biometric data, (vii) data concerning health, (viii) sex life or sexual orientation (Mondschein et al, 2019). 

Regarding \textbf{personality traits}, these could be traits from an established personality models/frameworks such as the ``Big Five'' or OCEAN model by Raymond Cattell and others. OCEAN defines five personality dimensions (Openness to Experience, Conscientiousness, Extraversion, Agreeableness and Neuroticism) (McCrae et al, 1992). An alternative framework to be used to research personality traits, would be the ``16-Personalities-Model'' from Myers-Briggs (EI/SN/TF/JP), also known as Myers-Briggs Type Indicator (MBTI). The MBTI categorizes individuals into 16 distinct personality types based on four dichotomies: Introversion vs. Extraversion, Sensing vs. Intuition, Thinking vs. Feeling, and Judging vs. Perceiving. These categories are meant to reflect an individual's preferences in how they perceive the world and make decisions. (Brown et al, 2009).

The literature review explores the question which standardised frameworks are used for capturing such personal attributes and attribute categories, or personal traits and personal traits categories and synthesises the results.

\section{Literature search}

The following working definitions of the key terms relating to the research question are introduced: ``LLM error'' and ``Human factors in detecting an LLM error''.

\subsection{Steps 1 and 2: Identify journals and databases}

Although vom Brocke et al (2009) recommends (but not mandates) to identify journals first and then query databases which index them second, for this review the steps have been reversed. The reason for this is two-fold:
Firstly, while the question of reducing LLM hallucinations with technical means (i.e. improving LLM model performance) is in the information systems domain, the questions of LLM usage (in general) and specifically the question of human factors in detecting LLM hallucinations is not, due to the broad applicability of LLMs in all professions. Even a first shallow literature screening brought up examples from many domains outside information systems domain, for example healthcare / medical sciences. A typical pre-selection of high-quality journals and conference proceedings from just one domain, such as the ``basket of eight'' in information systems / business informatics, would narrow the search too much.
Secondly, the broad societal use of LLMs only started with the ChatGPT breakthrough at the end of 2022, and the state-of-the-art LLM technology GPT has only been invented at the end of 2017. Therefore, the available time period for the systematic literature review is quite narrow (2018ff). Restricting the journals too much at first would yield too few results.

Therefore, the steps for identifying the journals are reversed as follows:
\begin{enumerate}
    \item Query one large scientific database, Scopus1, with the defined keywords related to the research question. 
    \item Restrict the number of journals/proceedings to include based on journal/proceedings reputation according to Scopus. Disregard journals/proceedings with insufficient reputation (as defined in step 4).
\end{enumerate}

\subsection{Step 3: Define search terms}

Based on the research question and topic conceptualisation, the search terms are defined as follows, leading to 15 search terms such as e.g. ``gpt failure'' or ``ChatGPT hallucination'':

\texttt{(("llm" OR "large language model" OR "gpt" OR "generative pretrained transformer" OR "ChatGPT") AND ("error" OR "failure" OR "hallucination"))}

Further configuration of Scopus search is the following.
\begin{itemize}
    \item The terms are searched in article title, abstract and keywords (standard config)
    \item Year range: 2018ff
    \item Document type: Article, Conference paper, Book chapter
    \item Language: English
    \item Source type: Journal, Conference proceeding, Book series
\end{itemize}

The full Scopus search configuration is replicated here for convenience:

\texttt{TITLE-ABS-KEY ( ( ( "llm"  OR  "large language model"  OR  "gpt"  OR  "generative pretrained transformer"  OR  "ChatGPT" )  AND  ( "error"  OR  "failure"  OR  "hallucination" ) ) )  AND  PUBYEAR  >  2017  AND  PUBYEAR  <  2025  AND  ( LIMIT-TO ( DOCTYPE ,  "ar" )  OR  LIMIT-TO ( DOCTYPE ,  "cp" )  OR  LIMIT-TO ( DOCTYPE ,  "ch" ) )  AND  ( LIMIT-TO ( LANGUAGE ,  "English" ) )  AND  ( LIMIT-TO ( SRCTYPE ,  "k" )  OR  LIMIT-TO ( SRCTYPE ,  "p" )  OR  LIMIT-TO ( SRCTYPE ,  "j" ) )}

\subsection{Step 4: Perform search and select sources and papers}

As of January 11th, 2024 (cutoff date for this work), this search yielded 594 results. Initial Scopus search results analysis shows that the topic was slowly growing in 2018-2021, and then exploded in 2023, showing that is a topic of hot research, coinciding with ChatGPT release and success. Breaking down the search results by subject area shows that the ``keywords first, journal selection second'' search strategy is justifiable. Only 30.6\% are in Computer Science, while the rest of results is spread over many other subject areas, the two largest after Computer Sciences being Social Sciences (11.0\%) and Medicine (10.6\%). Restricting the search strategy to ``journal first'' would restrict the results too much due to the broadness of the topic of Generative AI reaching beyond Computer Science / Business Informatics.

As a next step, the 594 search results are narrowed down based on analysing the title and abstract related to the research question and conceptual background (LLM errors, human factors in detecting these errors). During this task, for each selected document deemed relevant for further reading, the document journal’s SCIMago Journal Rank (SJR) and highest achieved ranking percentile is collected as a basis for subsequent filtering of search results to filter to a best possible mix of document and journal relevance for the research question.

\begin{center}
\begin{figure}[!h]
\begin{centering}
\includegraphics[width=11cm]{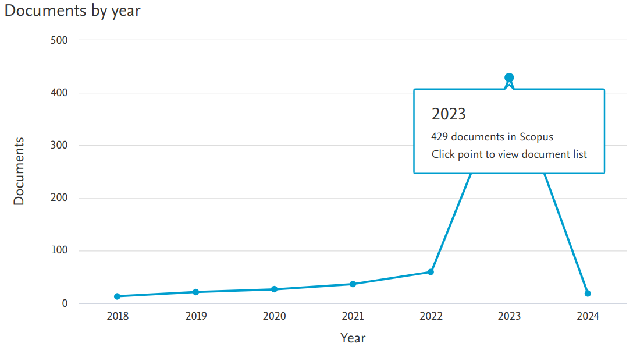}
\par\end{centering}
\caption{Scopus search results by year}
\end{figure}
\par\end{center}

\begin{center}
\begin{figure}[!h]
\begin{centering}
\includegraphics[width=11cm]{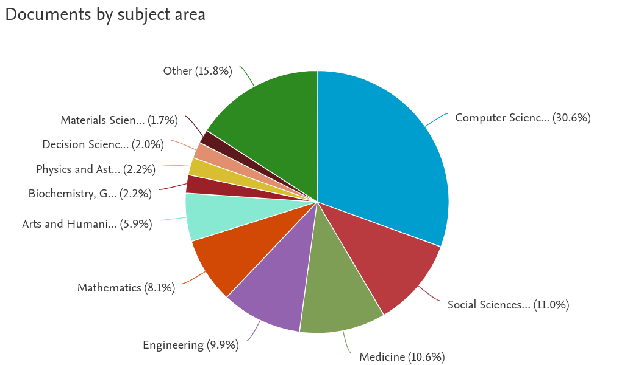}
\par\end{centering}
\caption{Scopus search results by subject area}
\end{figure}
\par\end{center}

After title \& abstract analysis, 61 documents remain from 51 journals. The full result list is documented in the appendix. For further analysis, the title selection is further reduced by selecting journals with a high reputation. This is done by filtering for journals whose ranking is at least in the 75th percentile (i.e. Upper Quartile) of all journals in their respective subject area. This results in a final list of 28 titles included for further analysis. The results table is sorted by the column ``Scopus top perc. 2022'' descending.

\begin{longtable}{|p{0.1cm}|p{4.5cm}|p{0.5cm}|p{2.5cm}|p{2cm}|p{0.9cm}|p{0.9cm}|}
\caption{Selected documents based on journal reputation $\geq$75th percentile (Upper Quartile)} \label{table2}\\
\hline
\textbf{\#} & \textbf{Title} & \textbf{Year} & \textbf{Journal} & \textbf{Publisher} & \textbf{Scopus SJR 2022} & \textbf{Scopus top perc. 2022} \\
\hline
\endfirsthead

\multicolumn{7}{c}%
{\tablename\ \thetable\ -- \textit{Continued from previous page}} \\
\hline
\textbf{\#} & \textbf{Title} & \textbf{Year} & \textbf{Journal} & \textbf{Publisher} & \textbf{Scopus SJR 2022} & \textbf{Scopus top perc. 2022} \\
\hline
\endhead

\hline \multicolumn{7}{r}{\textit{Continued on next page}} \\
\endfoot

\hline
\endlastfoot

1 & Benefits, Limits, and Risks of GPT-4 as an AI Chatbot for Medicine & 2023 & New England Journal of Medicine & Massachusetts Medical Society & 26,015 & 99 \\
2 & Evaluating large language models on medical evidence summarization & 2023 & npj Digital Medicine & Nature Research & 3,552 & 99 \\
3 & Artificial intelligence versus Maya Angelou: Experimental evidence that people cannot differentiate AI-generated from human-written poetry & 2021 & Computers in Human Behavior & Elsevier Ltd & 2,464 & 99 \\
4 & Believe in Artificial Intelligence? A User Study on the ChatGPT‘s Fake Information Impact & 2023 & IEEE Transactions on Computational Social Systems & Institute of Electrical and Electronics Engineers Inc. & 1,351 & 98 \\
5 & A GPT-4 Reticular Chemist for Guiding MOF Discovery** & 2023 & Angewandte Chemie - International Edition & John Wiley and Sons Inc & 5,573 & 96 \\
6 & Can ChatGPT Evaluate Plans? & 2023 & Journal of the American Planning Association & Routledge & 2,572 & 96 \\
7 & Thus spoke GPT-3: Interviewing a large-language model on climate finance & 2023 & Finance Research Letters & Elsevier Ltd & 2,231 & 96 \\
8 & Performance of ChatGPT, GPT-4, and Google Bard on a Neurosurgery Oral Boards Preparation Question Bank & 2023 & Neurosurgery & Wolters Kluwer Medknow Publications & 1,221 & 96 \\
9 & Exploring the Intersection of Artificial Intelligence and Neurosurgery: Let us be Cautious With ChatGPT & 2023 & Neurosurgery & Wolters Kluwer Medknow Publications & 1,221 & 96 \\
10 & Assessing student errors in experimentation using artificial intelligence and large language models: A comparative study with human raters & 2023 & Computers and Education: Artificial Intelligence & Elsevier B.V. & 1,700 & 95 \\
11 & Performance of Generative Large Language Models on Ophthalmology Board–Style Questions & 2023 & American Journal of Ophthalmology & Elsevier Inc. & 1,895 & 94 \\
12 & Appraising the performance of ChatGPT in psychiatry using 100 clinical case vignettes & 2023 & Asian Journal of Psychiatry & Elsevier B.V. & 1,326 & 94 \\
13 & Benchmarking large language models’ performances for myopia care: a comparative analysis of ChatGPT-3.5, ChatGPT-4.0, and Google Bard & 2023 & eBioMedicine & Elsevier B.V. & 2,900 & 93 \\
14 & Assessing the Utility of ChatGPT Throughout the Entire Clinical Workflow: Development and Usability Study & 2023 & Journal of Medical Internet Research & JMIR Publications Inc. & 1,992 & 93 \\
15 & Large Language Models for Therapy Recommendations Across 3 Clinical Specialties: Comparative Study & 2023 & Journal of Medical Internet Research & JMIR Publications Inc. & 1,992 & 93 \\
16 & Artificial Intelligence Can Generate Fraudulent but Authentic-Looking Scientific Medical Articles: Pandora's Box Has Been Opened & 2023 & Journal of Medical Internet Research & JMIR Publications Inc. & 1,992 & 93 \\
17 & Hallucination Detection: Robustly Discerning Reliable Answers in Large Language Models & 2023 & International Conference on Information and Knowledge Management, Proceedings & Association for Computing Machinery & 1,214 & 92 \\
18 & A pilot study on the efficacy of GPT-4 in providing orthopedic treatment recommendations from MRI reports & 2023 & Scientific Reports & Nature Research & 973 & 92 \\
19 & Exploring the Role of Artificial Intelligence Chatbots in Preoperative Counseling for Head and Neck Cancer Surgery & 2023 & Laryngoscope & John Wiley and Sons Inc & 1,103 & 91 \\
20 & Comparison of three chatbots as an assistant for problem-solving in clinical laboratory & 2023 & Clinical Chemistry and Laboratory Medicine & Walter de Gruyter GmbH & 1,266 & 88 \\
21 & Performance and exploration of ChatGPT in medical examination, records and education in Chinese: Pave the way for medical AI & 2023 & International Journal of Medical Informatics & Elsevier Ireland Ltd & 1,197 & 88 \\
22 & USE OF ARTIFICIAL INTELLIGENCE LARGE LANGUAGE MODELS AS A CLINICAL TOOL IN REHABILITATION MEDICINE: A COMPARATIVE TEST CASE & 2023 & Journal of Rehabilitation Medicine & Medical Journals Sweden AB & 887 & 88 \\
23 & Assessment of Resident and AI Chatbot Performance on the University of Toronto Family Medicine Residency Progress Test: Comparative Study & 2023 & JMIR Medical Education & JMIR Publications Inc. & 837 & 86 \\
24 & ChatClimate: Grounding conversational AI in climate science & 2023 & Communications Earth and Environment & Nature Publishing Group & 2,457 & 84 \\
25 & Evaluating the Diagnostic Accuracy and Management Recommendations of ChatGPT in Uveitis & 2023 & Ocular Immunology and Inflammation & Taylor and Francis Ltd. & 805 & 84 \\
26 & To what extent does ChatGPT understand genetics? & 2023 & Innovations in Education and Teaching International & Routledge & 708 & 83 \\
27 & The Use of ChatGPT to Assist in Diagnosing Glaucoma Based on Clinical Case Reports & 2023 & Ophthalmology and Therapy & Adis & 1,020 & 81 \\
28 & Appropriateness of premature ovarian insufficiency recommendations provided by ChatGPT & 2023 & Menopause & Wolters Kluwer Health & 869 & 79 \\

\end{longtable}

\section{Literature Analysis and Synthesis}

For analysis and synthesis of the selected 28 papers, the following logical approach to grouping and presenting the concepts is defined. In addition to this, the key finding from each paper is documented in the results table.

\begin{enumerate}
    \item LLM use cases in scope (results table columns: 1A,1B,1C)
    \begin{itemize}
        \item Information retrieval
        \item Information summarization
        \item Problem solving and decision support
    \end{itemize}
    \item LLM errors (results table columns: 2A,2B)
    \begin{itemize}
        \item Incorrect facts
        \item Missing correct facts
    \end{itemize}
    \item LLM error detection methods (results table columns: 3A,3B)
    \begin{itemize}
        \item Technical method
        \item Human-in-the-loop method
    \end{itemize}
    \item Human factors in error detection (results table columns: 4A,4B,4C)
    \begin{itemize}
        \item Personal attributes
        \item Personality traits
        \item Frameworks
    \end{itemize}
\end{enumerate}

\subsection{Analysis}

Due to ChatGPT’s ability to analyse any given PDF document, the option to employ ChatGPT for the analysis task was tested briefly. Several attempts with varying prompting techniques did not produce adequate results. As the example below shows, for given document 1, ChatGPT produced too generic content and hallucinated quotes which cannot be found in the document 1 itself. Therefore, for the time being, the systematic literature review process itself seems safe from being augmented by LLM systems; this probably would be an interesting research topic though.

\begin{center}
\begin{figure}[H]
\begin{centering}
\includegraphics[width=8cm]{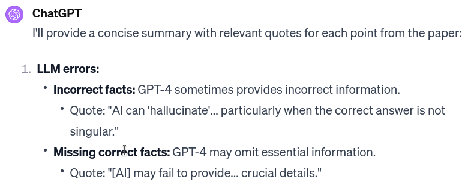}
\par\end{centering}
\caption{ChatGPT hallucinates quotes when analysing PDF documents from this research}
\end{figure}
\par\end{center}

During analysis, 7 of 28 papers had to be de-scoped due to low quality, or due to inaccessibility using the author’s institutional access (University of Duisburg-Essen). For the remaining 21 papers, the papers were analysed with the perspective of the concept matrix. An “X” indicates that a concept was in scope of the paper and notes for synthesis of concept across all papers could be taken. An “-” indicates that the concept was not in scope of the paper, which is also relevant for concept synthesis across all papers.

\newcolumntype{C}[1]{>{\centering\arraybackslash}p{#1}}
\newlength{\colwidth}
\setlength{\colwidth}{\dimexpr (\textwidth - 4pt*10)/11 \relax}

\begin{longtable}{|c|*{10}{C{\colwidth}|}}
\caption{Concept matrix from analysis of selected 28 documents} \label{table3}\\
\hline
\textbf{\#} & \textbf{1A} & \textbf{1B} & \textbf{1C} & \textbf{2A} & \textbf{2B} & \textbf{3A} & \textbf{3B} & \textbf{4A} & \textbf{4B} & \textbf{4C} \\
\hline
\endfirsthead

\multicolumn{11}{c}%
{{\bfseries \tablename\ \thetable{} -- continued from previous page}} \\
\hline 
\textbf{\#} & \textbf{1A} & \textbf{1B} & \textbf{1C} & \textbf{2A} & \textbf{2B} & \textbf{3A} & \textbf{3B} & \textbf{4A} & \textbf{4B} & \textbf{4C} \\\hline
\endhead

\hline \multicolumn{11}{|r|}{{Continued on next page}} \\ \hline
\endfoot

\hline
\endlastfoot

1 & X & X & X & X & X & - & X & - & X & - \\
\multicolumn{11}{|l|}{medical professionals should be ``careful'' and ``cautious'' to make best use of LLMs for medical assistance} \\
\hline
2 & - & X & - & X & X & X & X & X & - & - \\
\multicolumn{11}{|l|}{human evaluation is key as automatic metrics fail to reflect true summary quality, challenging LLMs} \\
\hline
3 & - & - & X & - & - & - & X & X & - & - \\
\multicolumn{11}{|l|}{behavioural science shows human involvement crucial in monitoring AI creativity and influencing reactions} \\
\hline
4 & - & - & X & X & - & - & X & X & - & - \\
\multicolumn{11}{|l|}{need for verifying chatbot accuracy, mitigating risks, and improving response validation methods} \\
\hline
5 & - & - & X & X & - & - & X & X & - & - \\
\multicolumn{11}{|l|}{developed framework emphasizes a symbiotic human-AI collaboration} \\
\hline
6 & - & - & X & X & X & - & X & X & - & - \\
\multicolumn{11}{|l|}{explore best prompting practices and derive guidelines to use ChatGPT more efficiently and effectively} \\
\hline
8 & X & - & X & X & X & - & X & X & X & - \\
\multicolumn{11}{|l|}{distinguishing knowledge from its application; quantifying hallucinations and validating LLMs is crucial} \\
\hline
12 & - & - & X & - & - & - & X & X & - & - \\
\multicolumn{11}{|l|}{Research gaps: Prove usefulness; moral and legal considerations} \\
\hline
13 & X & - & X & X & X & - & X & X & - & - \\
\multicolumn{11}{|l|}{robust design to minimise bias from human evaluators} \\
\hline
14 & X & - & X & X & X & - & X & - & - & - \\
\multicolumn{11}{|l|}{analysing ChatGPT’s accuracy at not just one step but rather throughout the entire clinical workflow} \\
\hline
15 & - & - & X & X & X & - & X & X & - & - \\
\multicolumn{11}{|l|}{more robust validation of the method requires the inclusion of a larger number of physician specialists} \\
\hline
16 & - & - & X & - & - & - & X & X & - & - \\
\multicolumn{11}{|l|}{expert readers identify semantic inaccuracies and errors} \\
\hline
17 & X & - & X & X & X & X & - & - & - & - \\
\multicolumn{11}{|l|}{we address LLM hallucination by proposing a robust discriminator, RelD, trained on the constructed RelQA dataset} \\
\hline
18 & - & X & X & X & X & - & X & X & - & - \\
\multicolumn{11}{|l|}{Expert surgeons rated the recommendations at least as ``good''} \\
\hline
19 & X & - & - & X & X & - & X & X & - & - \\
\multicolumn{11}{|l|}{LLM info matches web search in accuracy, thoroughness, and error as judged by surgeons} \\
\hline
21 & X & - & X & X & X & - & X & X & - & - \\
\multicolumn{11}{|l|}{poor performance in critical thinking and reasoning skills, such as case analyses and selection of treatment options} \\
\hline
22 & - & - & X & X & X & - & X & X & - & - \\
\multicolumn{11}{|l|}{emphasizes the significance of using specific and appropriate prompts for generating accurate outcomes} \\
\hline
23 & X & - & X & X & X & - & X & - & - & - \\
\multicolumn{11}{|l|}{it would be beneficial to appraise the suitability of various LLM evaluation frameworks} \\
\hline
24 & X & - & - & X & X & X & - & X & - & - \\
\multicolumn{11}{|l|}{hallucinations mitigated with domain-specific fine-tuning} \\
\hline
26 & X & - & X & X & X & - & - & - & - & - \\
\multicolumn{11}{|l|}{ChatGPT can pass a Ph.D entrance exam in medical genetics} \\
\hline
27 & - & - & X & X & X & - & X & X & - & - \\
\multicolumn{11}{|l|}{performance on par or better than senior ophthalmology residents} \\
\hline
7 & \multicolumn{10}{|l|}{out scoped due to too small scope (just one interview)} \\
\hline
9 & \multicolumn{10}{|l|}{out scoped due to lack of access rights} \\
\hline
20 & \multicolumn{10}{|l|}{out scoped due to lack of access rights} \\
\hline
25 & \multicolumn{10}{|l|}{out scoped due to lack of access rights} \\
\hline
28 & \multicolumn{10}{|l|}{out scoped due to lack of access rights} \\
\hline

\end{longtable}

\subsection{Synthesis}

For synthesis, the document references are indicated in brackets “[]” referencing the document number from the analysis results table (column \#).

\underline{1. LLM use cases}

For most papers (19 of 21), the most complex LLM use case was in focus, problem solving and decision support. Examples for this are taking drug-related treatment decisions [1], creations of psychiatric treatment plans [12] and passing complex medical exams [8]. Querying information from the LLM is always a part of this use case too, but only 4 papers distinguish this aspect specifically in their LLM testing methods. For example, paper [1] distinguishes between simply querying information about a drug and taking the decision to recommend the use of this drug for a specific situation. Three papers focus on the querying information use case only, such as providing pre-surgery information [19]. Three papers focus on the information summarization use case, for example summarizing medical evidence [2]. Most papers are from the healthcare/medicine domain (18 of 21). Only one paper each is from different domains: urban planning, chemistry, high school education and climate science.

\underline{2. LLM errors}

While 16 of 21 papers discuss the LLM hallucination problem, only 8 of 21 papers discuss the LLM omission problem. One paper [18] was a notable exception with detecting none of the two errors in their LLM evaluation. The synthesis of the LLM error discussions across papers reveals that LLM hallucinations are aligned to a variety of quality evaluation dimensions such as “Factual Consistency” [2], “Accuracy” [13], “Falseness” [15], “Correctness” [17]. Alignment of LLM omissions to a quality evaluation dimension across two papers ([2],[13]) is consistent with “Comprehensiveness”.
Paper [21] reveals a framework for further categorization of LLM errors, “open-vs.-closed-domain hallucination”, which informs the root cause of the hallucination as externally-induced (“open” - error in LLM training material) or not (“closed” – error related to incorrect LLM output based on correct external data).
The papers include a plethora of catchy examples for each error domain, useful for communicating the error types, such as:
Hallucinations: LLM states that it has a master’s degree [1]; makes errors in dosing of drugs [14]; produces non-existing citations [16]; predicts Scotland’s glaciers will melt due to climate change even though Scotland does not have any [24]
Omissions: in summarizing a patient/doctor conversation transcript omitted summarizing the ordered blood tests [1]; omitted mentioning of complications associated with refractive surgeries [13]; omits a vital second drug to accompany the first drug prescription [15]

\underline{3. LLM error detection methods}

Only four papers evaluate technical methods for detecting LLM errors which are potentially automatable. All papers describe various forms of human-in-the-loop methods for detecting LLM errors while evaluating LLM quality.
The discussed technical methods are the automated calculation of metrics ROUGE-L, METEOR and BLEU [2]. Paper [17] developed a machine learning model specialised for LLM error detection named RelD, trained on a specialised dataset RelQA curated for this purpose.
The human-in-the-loop LLM error detection methods all build up on the same generic process:
(1) Build a specialised LLM test dataset, or repurpose an existing test dataset such as an existing medical exam
(2) (optional) Build a specialised LLM prompting technique or repurpose an existing one
(3) Present the test dataset to the LLM to collect LLM responses, either in single Q\&A mode (e.g. answering a medical exam multiple choice question correctly), or in longer-lasting multi-step conversations
(4) Let humans rate the LLM responses from the perspective of various dimensions, both qualitatively (e.g. text report) and quantitatively (e.g. Likert scale)

\underline{4. Human factors in LLM error detection}

None of the analysed papers have a focus on researching the human factors influencing the ability to detect LLM errors. For the prevalent human-in-the-loop LLM error detection method used in most papers, research relies and builds upon involving those humans in the loop who have extensive domain expertise for the domain the LLMs are evaluated in, with a special emphasis on seniority. For example, paper [13] emphasizes the involvement of “highly experienced consultant-level paediatric ophthalmologists (>7 years of expertise)”, paper [18] emphasizes the use of “two board-certified and specialty-trained senior orthopaedic surgeons with ten and 12 years of clinical experience in orthopaedic and trauma surgery”, and papers [24],[27] emphasize “expert” and “senior” knowledge.
In the analysed research, the number of humans involved in evaluating the LLMs is very small, typically ranging from two to three. In case this small number of evaluators disagreed on the qualitative evaluation of an LLM output (e.g. one detected an LLM error while another didn’t), rather than focussing on why there is disagreement, current research consolidates or amalgamizes the disparate rating into a unified one, using various methods such as Consensus protocols or metrics from the field of social sciences such as Cohens Kappa, Fleiss Kappa or Gwet AC1.
Apart from general domain expertise and general seniority, no other personal properties (such as age or gender) of the human evaluators are disclosed or in focus of the research.
Concerning personality traits, paper [1] indicates an importance of “caution” and “carefulness” for interacting with LLMs, while paper [8] indicates an importance of “intuition” for those. Again, those traits are not in focus of the research, and thus no personality trait frameworks such as OCEAN or MBTI are used.
Summarizing the synthesis of this concept, the LLM quality evaluation (which includes LLM error identification) is in current research focus, rating the quality of the most used LLM ChatGPT or comparing ChatGPT quality against upcoming competitors such as Google Bard. Human factors involved in LLM error identification are currently not researched.

\section{Discussion}

The synthesis reveals that LLM usage, and thus also the shortcomings of LLM usage through occurring LLM errors, is currently most discussed in the healthcare/medicine domain, due to the reason that erroneous LLM-created medical advice could be potentially harmful to humans, when medical conditions are misdiagnosed or mistreated (Lee et al, 2023). This is especially relevant in high-stakes medical specialties like neurosurgery (Ali et al, 2023).
The synthesis reveals that LLM errors are in general a problem that needs solving and confirms the two prevalent types of LLM errors – hallucination of erroneous facts and omission of relevant facts, which are researched across all analysed papers. Mechanisms to handle hallucinations, omissions and errors should be incorporated into LLM systems (Lee et al, 2023). The synthesis indicates some current research bias towards discussion of the first LLM error type (hallucination – 16 papers) over the second (omission – 8 papers). Solving the problem of LLM errors by means of quality-testing LLM errors is an area of active research and increased importance, as the “exploding” number of papers on the subject since 2023 indicates. 
Currently, LLM quality is overwhelmingly tested by human-in-the-loop methods, even though older but automatable methods such as ROUGE-L, METEOR or BLEU exist (Tang et al, 2023). For the human-in-the-loop methods, a behavioural science approach is recommended to provide new insights into “human versus artificial intelligence” (Köbis et al, 2023). There is no off-the-shelf automatic evaluation metric specifically designed to assess the factuality of responses generated by LLM systems (Tang et al, 2023). A potential approach is to develop algorithms that can detect and automatically correct errors (Amaro et al, 2023). One such novel method is RelD, an ML model specialised for evaluating LLM quality, building upon a specialised RelQA dataset design (Chen et al, 2023). 

From the perspective of this literature review’s research question – “Which human factors influence the ability of humans to detect LLM system errors?” – the synthesis reveals that this is an aspect not in focus of current research. Current research relies on bringing few (up to ten) senior-/expert-level human domain experts into the human-in-the-loop testing of LLM quality. So, only the personal attributes “seniority” and “domain expertise/experience” are considered at all, without distinguishing different attributes or categories of these concepts (seniority, domain expertise). Other factors – personal attributes or personal traits – are not researched at all. 

\subsection{Future research agenda}

The future research agenda is structured by the concept matrix established for the analysis and synthesis.

\underline{1. LLM use case domains}

While the potential harmfulness is obvious in the healthcare/medical domain, in other domains it seems not well researched. For example, the long-term effects of erroneous urban planning due to LLM errors could be harmful to humans as well, on a different timescale yet even more impactful. The synthesis reveals a research bias towards the healthcare/medicine sector, leaving other subject areas under-researched.
This informs a potential research agenda to systematically evaluate the LLM quality and the impact of LLM errors in other use case domains than healthcare/medical.
Validation of LLMs on higher-order and open-ended scenarios is vital for the successful integration of these tools into high-stakes medical specialties (Ali et al, 2023). Currently LLMs perform poorly on clinical questions that evaluate critical thinking and reasoning skills, such as case analysis and treatment options (Wang et al, 2023). 
This informs a potential research agenda to systematically evaluate more complex task scenarios – in the medical use case domain or others – in the evaluation of LLM systems.

\underline{2. LLM error type}

LLM errors of type “omission” are under-researched. Omissions are relevant to use case domains with large corpuses of facts, such as requirements catalogues from law compliance, or complex fine-grained processes where omissions of process steps could be impactful or even harmful.
This informs a potential research agenda that those use case domains should be systematically identified and the LLM quality specifically measured regarding fact omissions.

\underline{3. LLM error detection method}

Most papers mention the necessity for mechanisms to manage LLM errors, emphasizing the importance of human oversight in monitoring and adjusting system outcomes. The papers highlight the inadequacy of current automatic metrics to gauge the quality of summaries and suggest the development of more effective evaluation methods, as well as the need for human evaluators to ensure accuracy and factuality. Furthermore, there's a call for research into more robust ways to verify chatbot responses, including the development of algorithms for automatic detection and correction of misinformation and the provision of user feedback mechanisms.
This informs a potential research agenda for both error detection solutions, technical and human-in-the-loop, with research gaps outlined in the next two paragraphs for each solution. An overarching research agenda would be appraising the suitability of various LLM evaluation frameworks (Huang et al, 2023).

\textit{Human-in-the-loop solutions}

The aspect of human trust in LLM systems is an area of potential research. To what extent can the user “trust” the LLM, or does the user need to spend time verifying the veracity of what the LLM writes? (Lee et al, 2023). 
For this, also the need for an increased number of humans involved in error detection research is seen, because current error detection research relies on very few human experts (in analysed papers, always only less than ten domain experts were involved).
Another area of research should be the exploration of best prompting practices and guideline derivation for more effective use of LLM systems (Fu et al, 2023). This would emphasize the significance of using specific and appropriate prompts for generating accurate outcomes (Zhang et al, 2023).

\textit{Technical solutions}

More effective automatic evaluation methods should be developed (Tang et al, 2023). From all analysed papers, only the RelD method is such a novel automatable method (Chen et al, 2023). To evolve this, new datasets from other use case domains would have to be created and evaluated, and the generalisability of the RelD approach evaluated.

\underline{4. Human factors in detecting LLM errors:}

An increased number of humans involved in the human-in-the-loop testing process is recommended to make research more robust (Wilhelm et al, 2023). Such studies should have a robust design to minimise bias from human evaluators (Lim et al, 2023). The impact of personal properties and/or personality traits on the human ability to detect LLM errors has not been researched yet and thus is a potential area of research.

\subsection{Limitations}

Due to given time and effort constraints for this work, a backward and forward literature search for the selected 28 papers was not performed. Only one scientific indexing database, Scopus, was consulted, albeit a large one.

\newpage{}

\section{Bibliography}

\raggedright

\textit{Literature: Chapters 1, 2, 3, 4}

Mollick, E. (2022). ChatGPT Is a Tipping Point for AI. Harvard Business Review Online. \url{https://hbr.org/2022/12/chatgpt-is-a-tipping-point-for-ai} 

Hu, K. (2023). ChatGPT sets record for fastest-growing user base - analyst note. Reuters. \url{https://www.reuters.com/technology/chatgpt-sets-record-fastest-growing-user-base-analyst-note-2023-02-01/}

Vaswani et al (2017). Attention is all you need. Advances in Neural Information Processing Systems. Volume 2017-December, Pages 5999 – 6009. 2017 31st Annual Conference on Neural Information Processing Systems, NIPS 2017. \url{https://proceedings.neurips.cc/paper/7181-attention-is-all}

vom Brocke et al (2009). RECONSTRUCTING THE GIANT: ON THE IMPORTANCE OF RIGOUR IN DOCUMENTING THE LITERATURE SEARCH PROCESS. ECIS 2009 Proceedings. 161. \url{https://aisel.aisnet.org/ecis2009/161}
  
Hart, C. (1998). Doing a literature review: releasing the social science research imagination. Sage Publications, London.

Cooper, H.M. (1988). Organizing knowledge syntheses: A taxonomy of literature reviews. Knowledge in Society 1, 104. \url{https://doi.org/10.1007/BF03177550}
  
Webster, J., \& Watson, R. T. (2002). Analyzing the Past to Prepare for the Future: Writing a Literature Review. MIS Quarterly, 26(2), xiii–xxiii. \url{http://www.jstor.org/stable/4132319}
  
Eloundou et al (2023). GPTs are GPTs: An Early Look at the Labor Market Impact Potential of Large Language Models. ArXiv Preprint. \url{https://arxiv.org/abs/2303.10130}
  
Maynez et al (2020). On Faithfulness and Factuality in Abstractive Summarization. In Proceedings of the 58th Annual Meeting of the Association for Computational Linguistics, pages 1906–1919, Online. Association for Computational Linguistics. \url{https://aclanthology.org/2020.acl-main.173/}
  
Ji et al (2023). Survey of Hallucination in Natural Language Generation. ACM Computing Surveys. Volume 55. Issue 12. Article No.: 248 pp 1–38. \url{https://doi.org/10.1145/3571730}
  
Emsley, R. (2023). ChatGPT: these are not hallucinations – they’re fabrications and falsifications. Schizophr 9, 52 (2023). \url{https://doi.org/10.1038/s41537-023-00379-4}

McCrae, R. et al (1992). An introduction to the five-factor model and its applications. Journal of Personality, 60(2), 175–215. \url{https://doi.org/10.1111/j.1467-6494.1992}
 
Brown, F. et al (2009). The Myers-Briggs type indicator and transformational leadership. Journal of Management Development, 28(10), 916-932. \url{https://doi.org/10.1108/02621710911000677}
 
Mondschein, C. et al (2019). The EU’s General Data Protection Regulation (GDPR) in a Research Context. Chapter 5 of “Fundamentals of Clinical Data Science”. Springer Open. \url{https://doi.org/10.1007/978-3-319-99713-1} 

\textit{Literature: Chapters 5, 6 (21 papers analysed)}

Lee, P. et al (2023). Benefits, Limits, and Risks of GPT-4 as an AI Chatbot for Medicine. The New England Journal of Medicine. 388;13. \url{https://doi.org/10.1056/NEJMsr2214184}
 
Tang, L. et al (2023). Evaluating large language models on medical evidence Summarization. Nature npj Digital Medicine 6:158. \url{https://doi.org/10.1038/s41746-023-00896-7}
 
Köbis, N. et al (2021). Artificial intelligence versus Maya Angelou: Experimental evidence that people cannot differentiate AI-generated from human-written poetry. Computers in Human Behavior 114 (2021) 106553. \url{https://doi.org/10.1016/j.chb.2020.106553}
 
Amaro, I. et al (2023). Believe in Artificial Intelligence? A User Study on the ChatGPT’s Fake Information Impact. IEEE TRANSACTIONS ON COMPUTATIONAL SOCIAL SYSTEMS. \url{https://doi.org/10.1109/TCSS.2023.3291539}
 
Zheng, Z. et al (2023). A GPT-4 Reticular Chemist for Guiding MOF Discovery. Angew. Chem. Int. Ed. 2023, 62, e202311983. \url{https://doi.org/10.1002/anie.202311983}
 
Fu, X. et al (2023). Can ChatGPT Evaluate Plans? Journal of the American Planning Association. \url{https://doi.org/10.1080/01944363.2023.2271893} 

Ali, R. et al (2023). Performance of ChatGPT, GPT-4, and Google Bard on a Neurosurgery Oral Boards Preparation Question Bank. Congress of Neurological Surgeons, VOLUME 93 | NUMBER 5. \url{https://doi.org/10.1227/neu.0000000000002551}
 
D’Souza, R. et al (2023). Appraising the performance of ChatGPT in psychiatry using 100 clinical case vignettes. Asian Journal of Psychiatry 89 (2023) 103770. \url{https://doi.org/10.1016/j.ajp.2023.103770}
 
Lim, Z. et al (2023). Benchmarking large language models’ performances for myopia care: a comparative analysis of ChatGPT-3.5, ChatGPT-4.0, and Google Bard. eBioMedicine 2023;95: 104770. \url{https://doi.org/10.1016/j.ebiom.2023.104770}
 
Rao, A. et al (2023). Assessing the Utility of ChatGPT Throughout the Entire Clinical Workflow: Development and Usability Study. J Med Internet Res 2023 | vol. 25 | e48659. \url{https://doi.org/10.2196/48659}
 
Wilhelm, T. et al (2023). Large Language Models for Therapy Recommendations Across 3 Clinical Specialties: Comparative Study. J Med Internet Res 2023 | vol. 25 | e49324. \url{https://doi.org/10.2196/49324}
 
Májovský, M. et al (2023). Artificial Intelligence Can Generate Fraudulent but Authentic-Looking Scientific Medical Articles: Pandora’s Box Has Been Opened. J Med Internet Res 2023 | vol. 25 | e46924. \url{https://doi.org/10.2196/46924}
 
Chen, Y. et al (2023). Hallucination Detection: Robustly Discerning Reliable Answers in Large Language Models. CIKM ’23, October 21–25, 2023, Birmingham, United Kingdom. \url{https://doi.org/10.1145/3583780.3614905}
 
Truhn, D. et al (2023). A pilot study on the efficacy of GPT‑4 in providing orthopedic treatment recommendations from MRI reports. Nature Scientific Reports. (2023) 13:20159. \url{https://doi.org/10.1038/s41598-023-47500-2}
 
Lee, J. et al (2023). Exploring the Role of Artificial Intelligence Chatbots in Preoperative Counseling for Head and Neck Cancer Surgery. Laryngoscope, 00:1–5, 2023. \url{https://doi.org/10.1002/lary.31243}
 
Wang, H. et al (2023). Performance and exploration of ChatGPT in medical examination, records and education in Chinese: Pave the way for medical AI. International Journal of Medical Informatics 177 (2023) 105173. \url{https://doi.org/10.1016/j.ijmedinf.2023.105173}
 
Zhang, L. et al (2023). USE OF ARTIFICIAL INTELLIGENCE LARGE LANGUAGE MODELS AS A CLINICAL TOOL IN REHABILITATION MEDICINE: A COMPARATIVE TEST CASE. J Rehabil Med 2023; 55: jrm13373. \url{https://doi.org/10.2340/jrm.v55.13373}
 
Huang, R. et al (2023). Assessment of Resident and AI Chatbot Performance on the University of Toronto Family Medicine Residency Progress Test: Comparative Study. JMIR Med Educ 2023 | vol. 9 | e50514. \url{https://doi.org/10.2196/50514}
 
Vaghefi, S. et al (2023). ChatClimate: Grounding conversational AI in climate science. Nature COMMUNICATIONS EARTH \& ENVIRONMENT | (2023) 4:480. \url{https://doi.org/10.1038/s43247-023-01084-x}
 
Khosravi, T. et al (2023). To what extent does ChatGPT understand genetics? Innovations in Education and Teaching International. \url{https://doi.org/10.1080/14703297.2023.2258842}
 
Delsoz, M. et al (2023). The Use of ChatGPT to Assist in Diagnosing Glaucoma Based on Clinical Case Reports. Ophthalmol Ther (2023) 12:3121–3132. \url{https://doi.org/10.1007/s40123-023-00805-x} 

\newpage{}

\section{Appendix A – Full results of literature search}

"RR" in column 1 = "Relevance Ranking" according to Scopus results lists.

\begin{longtable}{|p{0.3cm}|p{4.5cm}|p{0.5cm}|p{2.5cm}|p{2cm}|p{0.9cm}|p{0.9cm}|}
\caption{Full results of literature search} \label{table4}\\
\hline
\textbf{RR} & \textbf{Title} & \textbf{Year} & \textbf{Journal} & \textbf{Publisher} & \textbf{Scopus SJR 2022} & \textbf{Scopus top perc. 2022} \\
\hline
\endfirsthead

\multicolumn{7}{c}%
{\tablename\ \thetable\ -- \textit{Continued from previous page}} \\
\hline
\textbf{RR} & \textbf{Title} & \textbf{Year} & \textbf{Journal} & \textbf{Publisher} & \textbf{Scopus SJR 2022} & \textbf{Scopus top perc. 2022} \\
\hline
\endhead

\hline \multicolumn{7}{r}{\textit{Continued on next page}} \\
\endfoot

\hline
\endlastfoot

18 & Benefits, Limits, and Risks of GPT-4 as an AI Chatbot for Medicine & 2023 & New England Journal of Medicine & Massachusetts Medical Society & 26,015 & 99 \\
24 & Evaluating large language models on medical evidence summarization & 2023 & npj Digital Medicine & Nature Research & 3,552 & 99 \\
58 & Artificial intelligence versus Maya Angelou: Experimental evidence that people cannot differentiate AI-generated from human-written poetry & 2021 & Computers in Human Behavior & Elsevier Ltd & 2,464 & 99 \\
53 & Believe in Artificial Intelligence? A User Study on the ChatGPT's Fake Information Impact & 2023 & IEEE Transactions on Computational Social Systems & Institute of Electrical and Electronics Engineers Inc. & 1,351 & 98 \\
34 & A GPT-4 Reticular Chemist for Guiding MOF Discovery** & 2023 & Angewandte Chemie - International Edition & John Wiley and Sons Inc & 5,573 & 96 \\
16 & Can ChatGPT Evaluate Plans? & 2023 & Journal of the American Planning Association & Routledge & 2,572 & 96 \\
6 & Thus spoke GPT-3: Interviewing a large-language model on climate finance & 2023 & Finance Research Letters & Elsevier Ltd & 2,231 & 96 \\
11 & Performance of ChatGPT, GPT-4, and Google Bard on a Neurosurgery Oral Boards Preparation Question Bank & 2023 & Neurosurgery & Wolters Kluwer Medknow Publications & 1,221 & 96 \\
49 & Exploring the Intersection of Artificial Intelligence and Neurosurgery: Let us be Cautious With ChatGPT & 2023 & Neurosurgery & Wolters Kluwer Medknow Publications & 1,221 & 96 \\
31 & Assessing student errors in experimentation using artificial intelligence and large language models: A comparative study with human raters & 2023 & Computers and Education: Artificial Intelligence & Elsevier B.V. & 1,700 & 95 \\
4 & Performance of Generative Large Language Models on Ophthalmology Board–Style Questions & 2023 & American Journal of Ophthalmology & Elsevier Inc. & 1,895 & 94 \\
54 & Appraising the performance of ChatGPT in psychiatry using 100 clinical case vignettes & 2023 & Asian Journal of Psychiatry & Elsevier B.V. & 1,326 & 94 \\
14 & Benchmarking large language models’ performances for myopia care: a comparative analysis of ChatGPT-3.5, ChatGPT-4.0, and Google Bard & 2023 & eBioMedicine & Elsevier B.V. & 2,900 & 93 \\
15 & Assessing the Utility of ChatGPT Throughout the Entire Clinical Workflow: Development and Usability Study & 2023 & Journal of Medical Internet Research & JMIR Publications Inc. & 1,992 & 93 \\
30 & Large Language Models for Therapy Recommendations Across 3 Clinical Specialties: Comparative Study & 2023 & Journal of Medical Internet Research & JMIR Publications Inc. & 1,992 & 93 \\
51 & Artificial Intelligence Can Generate Fraudulent but Authentic-Looking Scientific Medical Articles: Pandora's Box Has Been Opened & 2023 & Journal of Medical Internet Research & JMIR Publications Inc. & 1,992 & 93 \\
35 & Hallucination Detection: Robustly Discerning Reliable Answers in Large Language Models & 2023 & International Conference on Information and Knowledge Management, Proceedings & Association for Computing Machinery & 1,214 & 92 \\
23 & A pilot study on the efficacy of GPT-4 in providing orthopedic treatment recommendations from MRI reports & 2023 & Scientific Reports & Nature Research & 0,973 & 92 \\
48 & Exploring the Role of Artificial Intelligence Chatbots in Preoperative Counseling for Head and Neck Cancer Surgery & 2023 & Laryngoscope & John Wiley and Sons Inc & 1,103 & 91 \\
52 & Comparison of three chatbots as an assistant for problem-solving in clinical laboratory & 2023 & Clinical Chemistry and Laboratory Medicine & Walter de Gruyter GmbH & 1,266 & 88 \\
22 & Performance and exploration of ChatGPT in medical examination, records and education in Chinese: Pave the way for medical AI & 2023 & International Journal of Medical Informatics & Elsevier Ireland Ltd & 1,197 & 88 \\
40 & USE OF ARTIFICIAL INTELLIGENCE LARGE LANGUAGE MODELS AS A CLINICAL TOOL IN REHABILITATION MEDICINE: A COMPARATIVE TEST CASE & 2023 & Journal of Rehabilitation Medicine & Medical Journals Sweden AB & 0,887 & 88 \\
17 & Assessment of Resident and AI Chatbot Performance on the University of Toronto Family Medicine Residency Progress Test: Comparative Study & 2023 & JMIR Medical Education & JMIR Publications Inc. & 0,837 & 86 \\
38 & ChatClimate: Grounding conversational AI in climate science & 2023 & Communications Earth and Environment & Nature Publishing Group & 2,457 & 84 \\
55 & Evaluating the Diagnostic Accuracy and Management Recommendations of ChatGPT in Uveitis & 2023 & Ocular Immunology and Inflammation & Taylor and Francis Ltd. & 0,805 & 84 \\
19 & To what extent does ChatGPT understand genetics? & 2023 & Innovations in Education and Teaching International & Routledge & 0,708 & 83 \\
3 & The Use of ChatGPT to Assist in Diagnosing Glaucoma Based on Clinical Case Reports & 2023 & Ophthalmology and Therapy & Adis & 1,020 & 81 \\
47 & Appropriateness of premature ovarian insufficiency recommendations provided by ChatGPT & 2023 & Menopause & Wolters Kluwer Health & 0,869 & 79 \\
50 & OpenAI's GPT-4 performs to a high degree on board-style dermatology questions & 2024 & International Journal of Dermatology & John Wiley and Sons Inc & 0,623 & 74 \\
44 & Diagnostic and Management Performance of ChatGPT in Obstetrics and Gynecology & 2023 & Gynecologic and Obstetric Investigation & S. Karger AG & 0,606 & 74 \\
61 & The impact of human-AI collaboration types on consumer evaluation and usage intention: a perspective of responsibility attribution & 2023 & Frontiers in Psychology & Frontiers Media SA & 0,891 & 73 \\
12 & A Culturally Sensitive Test to Evaluate Nuanced GPT Hallucination & 2023 & IEEE Transactions on Artificial Intelligence & Institute of Electrical and Electronics Engineers Inc. & 1,381 & 66 \\
2 & Generative artificial intelligence fails to provide sufficiently accurate recommendations when compared to established breast reconstruction surgery guidelines & 2023 & Journal of Plastic, Reconstructive and Aesthetic Surgery & Churchill Livingstone & 0,826 & 64 \\
45 & GPT-4 Underperforms Experts in Detecting IV Fluid Contamination & 2023 & Journal of Applied Laboratory Medicine & Oxford University Press & 0,520 & 64 \\
9 & Comparative study of ChatGPT and human evaluators on the assessment of medical literature according to recognised reporting standards & 2023 & BMJ Health and Care Informatics & BMJ Publishing Group & 0,871 & 63 \\
28 & Comparison of Patient Education Materials Generated by Chat Generative Pre-Trained Transformer Versus Experts: An Innovative Way to Increase Readability of Patient Education Materials & 2023 & Annals of Plastic Surgery &  & 0,608 & 57 \\
43 & ChatGPT and Lacrimal Drainage Disorders: Performance and Scope of Improvement & 2023 & Ophthalmic Plastic and Reconstructive Surgery & Wolters Kluwer Health & 0,504 & 53 \\
36 & Evaluating the Current Ability of ChatGPT to Assist in Professional Otolaryngology Education & 2023 & OTO Open &  & 0,585 & 48 \\
21 & Knowledge Injection to Counter Large Language Model (LLM) Hallucination & 2023 & Lecture Notes in Computer Science (including subseries Lecture Notes in Artificial Intelligence and Lecture Notes in Bioinformatics) & Springer Science and Business Media Deutschland GmbH & 0,320 & 46 \\
33 & CORE-GPT: Combining Open Access Research and Large Language Models for Credible, Trustworthy Question Answering & 2023 & Lecture Notes in Computer Science (including subseries Lecture Notes in Artificial Intelligence and Lecture Notes in Bioinformatics) &  & 0,320 & 46 \\
39 & Roe: A Computational-Efficient Anti-hallucination Fine-Tuning Technology for Large Language Model Inspired by Human Learning Process & 2023 & Lecture Notes in Computer Science (including subseries Lecture Notes in Artificial Intelligence and Lecture Notes in Bioinformatics) &  & 0,320 & 46 \\
10 & AI chatbots not yet ready for clinical use & 2023 & Frontiers in Digital Health & Frontiers Media S.A. & 0,624 & 45 \\
56 & How Useful Are Educational Questions Generated by Large Language Models? & 2023 & Communications in Computer and Information Science & Springer Science and Business Media Deutschland GmbH & 0,194 & 34 \\
29 & Answering Natural Language Questions with OpenAI's GPT in the Petroleum Industry & 2023 & Proceedings - SPE Annual Technical Conference and Exhibition &  & 0,288 & 31 \\
1 & Generating multiple choice questions from a textbook: LLMs match human performance on most metrics & 2023 & CEUR Workshop Proceedings & CEUR-WS & 0,202 & 21 \\
5 & Gracenote.ai: Legal Generative AI for Regulatory Compliance & 2023 & CEUR Workshop Proceedings & CEUR-WS & 0,202 & 21 \\
25 & How Ready are Pre-trained Abstractive Models and LLMs for Legal Case Judgement Summarization? & 2023 & CEUR Workshop Proceedings & CEUR-WS & 0,202 & 21 \\
26 & Using ChatGPT for the FOIA Exemption 5 Deliberative Process Privilege & 2023 & CEUR Workshop Proceedings & CEUR-WS & 0,202 & 21 \\
32 & Do We Need Subject Matter Experts? A Case Study of Measuring Up GPT-4 Against Scholars in Topic Evaluation & 2023 & CEUR Workshop Proceedings &  & 0,202 & 21 \\
41 & Insights into Classifying and Mitigating LLMs' Hallucinations & 2023 & CEUR Workshop Proceedings &  & 0,202 & 21 \\
57 & Adapting Abstractive Summarization to Court Examinations in a Zero-Shot Setting: A Short Technical Paper & 2023 & CEUR Workshop Proceedings & CEUR-WS & 0,202 & 21 \\
7 & FROM TEXT TO DIAGNOSE: CHATGPT'S EFFICACY IN MEDICAL DECISION-MAKING & 2023 & Wiadomosci lekarskie (Warsaw, Poland : 1960) &  & 0,138 & 21 \\
8 & From ChatGPT to FactGPT: A Participatory Design Study to Mitigate the Effects of Large Language Model Hallucinations on Users & 2023 & ACM International Conference Proceeding Series & Association for Computing Machinery & 0,209 & 14 \\
20 & ChatGPT and Large Language Models in Healthcare: Opportunities and Risks & 2023 & 2023 IEEE International Conference on Artificial Intelligence, Blockchain, and Internet of Things, AIBThings 2023 - Proceedings & Institute of Electrical and Electronics Engineers Inc. & 0,000 & 0 \\
27 & Retrieval-Augmented Large Language Models for Adolescent Idiopathic Scoliosis Patients in Shared Decision-Making & 2023 & ACM-BCB 2023 - 14th ACM Conference on Bioinformatics, Computational Biology, and Health Informatics &  & 0,000 & 0 \\
46 & Capturing Failures of Large Language Models via Human Cognitive Biases & 2022 & Advances in Neural Information Processing Systems & Neural information processing systems foundation & 0,000 & 0 \\
42 & Supporting Human-AI Collaboration in Auditing LLMs with LLMs & 2023 & AIES 2023 - Proceedings of the 2023 AAAI/ACM Conference on AI, Ethics, and Society &  & 0,000 & 0 \\
13 & Dehallucinating Large Language Models Using Formal Methods Guided Iterative Prompting & 2023 & Proceedings - 2023 IEEE International Conference on Assured Autonomy, ICAA 2023 & Institute of Electrical and Electronics Engineers Inc. & 0,000 & 0 \\
60 & Real or Fake Text?: Investigating Human Ability to Detect Boundaries between Human-Written and Machine-Generated Text & 2023 & Proceedings of the 37th AAAI Conference on Artificial Intelligence, AAAI 2023 & AAAI Press & 0,000 & 0 \\
37 & Evaluating Open-Domain Question Answering in the Era of Large Language Models & 2023 & Proceedings of the Annual Meeting of the Association for Computational Linguistics &  & 0,000 & 0 \\
59 & A cross-sectional study to assess response generated by ChatGPT and ChatSonic to patient queries about Epilepsy & 2024 & Telematics and Informatics Reports & Elsevier B.V. & 0,000 & 0 \\


\end{longtable}

\end{document}